\documentclass[review,11pt]{elsarticle}

\usepackage{xcolor}
\usepackage{amssymb}
\usepackage{graphicx}
\usepackage{amssymb}
\usepackage{lscape}
\usepackage{multirow}
\usepackage[usenames,dvipsnames]{pstricks}
\usepackage{epsfig}
\usepackage{pst-grad} 
\usepackage{pst-plot} 
\usepackage{float}

\usepackage{amssymb,amsmath}
\usepackage{algorithm,algorithmic}
\usepackage{colortbl}
\usepackage{mathptmx}       
\usepackage{helvet}         
\usepackage{courier}        
\usepackage{type1cm}        
\usepackage[babel=true]{csquotes}               
\usepackage{subfigure,graphicx}
\usepackage{graphics}
\usepackage{subfigure,graphicx}
\usepackage{multicol}        
\usepackage[bottom]{footmisc}
\usepackage{ulem}
 \usepackage{soul}

\newcommand{\BetP}{\mathrm{BetP}}
\newcommand{\Id}{\mathrm{I_d}}

\DeclareMathOperator{\ocap}{\displaystyle{\small{\textcircled{{\scriptsize $\cap$}}}}}
\DeclareMathOperator{\owedge}{\displaystyle{\small{\textcircled{\scriptsize{$\wedge$}}}}}
\definecolor{light-gary}{gray}{0.95}
\definecolor{lgray}{gray}{0.85}
\begin{document}

\begin{frontmatter}



\title{Combining partially independent belief functions}


\author[corref,focal,rvt]{Mouna Chebbah}
\ead{Mouna.Chebbah@univ-rennes1.fr}

\author[rvt]{Arnaud Martin}
\ead{Arnaud.Martin@univ-rennes1.fr}

\author[els]{Boutheina Ben Yaghlane}
\ead{boutheina.yaghlane@ihec.rnu.tn}

\address[focal]{LARODEC Laboratory, University of Tunis, ISG Tunis, Tunisia}
\address[rvt]{IRISA, University of Rennes1, Lannion, France}
\address[els]{LARODEC Laboratory, University of Carthage, IHEC Carthage, Tunisia}

\begin{abstract}
The theory of belief functions manages uncertainty and also proposes a set of combination rules to aggregate opinions of several sources. Some combination rules mix evidential information where sources are independent; other rules are suited to combine evidential information held by dependent sources. In this paper we have two main contributions: First we suggest a method to quantify sources' degree of independence that may guide the choice of the more appropriate set of combination rules. Second, we propose a new combination rule that takes consideration of sources' degree of independence. The proposed method is illustrated on generated mass functions.
\end{abstract}


\end{frontmatter}


\section{Introduction}
Uncertainty theories like the \textit{theory of probabilities}, the \textit{theory of fuzzy sets} \cite{Zadeh65a}, the \textit{theory of possibilities} \cite{Dubois88} and the \textit{theory of belief functions} \cite{Dempster67a,Shafer76} model and manage uncertain data. The theory of belief functions can deal with imprecise and/or uncertain data provided by several belief holders and also combine them.\\
\indent Combining several evidential information held by distinct belief holders aggregates their points of view by stressing common points. In the theory of belief functions, many combination rules are proposed, some of them like \cite{Dubois88,Martin07b,Murphy00a,Smets94a,Yager87,Lefevre13a} are fitted to the aggregation of evidential information provided by cognitively independent sources whereas the \textit{cautious}, \textit{bold} \cite{Denoeux08a} and \textit{mean} combination rules can be applied when sources are \textit{cognitively dependent}. The choice of combination rules depends on sources independence.\\
\indent Some researches are focused on doxastic independence of variables such as \cite{BenYaghlane02a,BenYaghlane02b}; others \cite{Shafer76,Smets93a} tackled cognitive and evidential independence of variables. This paper is focused on measuring the independence of sources and not that of variables. We suggest a statistical approach to estimate the independence of sources on the bases of all evidential information that they provide. The aim of estimating the independence of sources is to guide the choice of the combination rule to be used when combining their evidential information.\\
\indent We propose also a new combination rule to aggregate evidential information and take into account the independence degree of their sources. The proposed combination rule is weighted with that degree of independence leading to the conjunctive rule \cite{Smets90a} when sources are fully independent and to the cautious rule \cite{Denoeux08a} when they are fully dependent.\\
\indent In the sequel, we introduce in Section 2 preliminaries of the theory of belief functions. In the Section 3, an evidential clustering algorithm is detailed. This clustering algorithm will be used in the first step of the independence measure process. Independence measure is then detailed in Section 4. It is estimated in four steps: In the first step the clustering algorithm is applied. Second a mapping between clusters is performed; then independence of clusters and sources are deduced in the last two steps. Independence is learned for only two sources and then generalized for a greater number of sources. A new combination rule is proposed in the Section 5 taking into account the independence degree of sources. The proposed method is tested on random mass functions in Section 6. Finally, conclusions are drawn. 
\section{Theory of belief functions}
\label{BFT}
The theory of belief functions was introduced by Dempster \cite{Dempster67a} and formalized by Shafer \cite{Shafer76} to model imperfect data. The \textit{frame of discernment} also called \textit{universe of discourse}, 
$\Omega=\{\omega_1,\omega_2, \ldots, \omega_N\}$, is an exhaustive set of $N$ mutually exclusive hypotheses $\omega_i$. The \textit{power set} $2^\Omega$ is a set of all subsets of $\Omega$; it is made of hypotheses and unions of hypotheses from $\Omega$. The \textit{basic belief assignment (\textsc{bba})} commonly called \textit{mass function} is a function defined on the power set $2^\Omega$ and spans the interval $[0,1]$ such that: 
\begin{equation}
\displaystyle{\sum_{A\subseteq \Omega}}m(A)=1
\label{eq2}
\end{equation}
A \textit{basic belief mass (\textsc{bbm})} also called \textit{mass}, $m(A)$, is a degree of faith on the truth of $A$. The \textsc{bbm}, $m(A)$, is a degree of belief on $A$ which can be committed to its subsets if further information justifies it \cite{Smets94a}.\\
\indent Subsets $A$ having a strictly positive mass are called \textit{focal elements}. Union of all focal elements is called \textit{core}. Shafer \cite{Shafer76} assumed a normality condition such that $m(\emptyset)=0$, thereafter Smets \cite{Smets90a} relaxed this condition in order to tolerate $m(\emptyset)>0$.\\
\indent The frame of discernment can also be a focal element; its \textsc{bbm}, $m(\Omega)$, is interpreted as a degree of \textit{ignorance}. In the case of \textit{total ignorance}, $m(\Omega)=1$.\\
\indent A \textit{simple support function} is a mass function with two focal elements including the frame of discernment. A simple support function $m$ is defined as follows:
\begin{equation}
m(A)=\left\{
\begin{tabular}{ll}
$1-w$&$if\hspace{0.1cm}A=B\hspace{0.1cm}\text{for some}\hspace{0.1cm}B\subset\Omega$\\
$w$&$\text{if}\hspace{0.1cm}A=\Omega$\\
$0$&$\text{otherwise}$\\
\end{tabular}
\right.
\end{equation}
Where $A$ is a \textit{focus} of that simple support function and $w\in[0,1]$ is its \textit{weight}. A simple support function is simply noted $A^{w}$. A nondogmatic mass function can be obtained by the combination of several simple support functions. Therefore, any nondogmatic mass function can be decomposed into several support functions using the \textit{canonical decomposition} proposed by Smets \cite{Smets95a}. \\
\indent The \textit{belief function} ($bel$) is computed from a \textsc{bba} $m$. The amount $bel(A)$ is the minimal belief on $A$ justified by available information on $B$ ($B\subseteq A$):
\begin{equation}
\label {bel}
bel(A)=\displaystyle{\sum_{B\subseteq A,B\neq\emptyset}}m(B)
\end{equation}

The \textit{plausibility function} ($pl$) is also derived from a \textsc{bba} $m$. The amount $pl(A)$ is the maximal belief on $A$ justified by information on $B$ which are not contradictory with $A$ ($A\cap B \neq\emptyset$):
\begin{equation}
\label{pl}
pl(A)=\displaystyle{\sum_{A\cap B\neq\emptyset}}m(B)
\end{equation}

Pignistic transformation computes pignistic probabilities from mass functions in the purpose of making a decision. The pignistic probability of a single hypothesis $A$ is given by:
\begin{eqnarray}
\label{pignistic}
BetP(A)=\sum_{B \subseteq \Omega, B \neq \emptyset} \frac{|B \cap A|}{|B|} \frac{m(B)}{1-m(\emptyset)}.
\end{eqnarray}
Decision is made according to the maximum pignistic probability. The single point having the greatest $BetP$ is the most likely hypothesis.
\subsection{Discounting}
Sources of information are not always reliable, they can be unreliable or even a little bit reliable. Taking into account reliability of sources, we adjust their beliefs proportionally to degrees of reliability.
Discounting mass functions is a way of taking consideration of sources' reliabilities into their mass functions. If reliability rate $\alpha$ of a source is known or can be quantified; discounting its mass function $m$ is defined as follows:
 \begin{equation}
\label{Equation(9)}
\left\{
\begin{tabular}{lll}
$m^{\alpha}(A)$&$=\alpha\times m(A)$&$,\forall A\subset\Omega$\\
$m^{\alpha}(\Omega)$&$=1-\alpha\times (1-m(\Omega))$\\
\end{tabular}
\right.
\end{equation}
This discounting operator can be used not only to take consideration of source's reliability, but also to consider any information which can be integrated into the mass function, $(1-\alpha)$ is called \textit{discounting rate}.
\subsection{Combination rules}
\label{CR}
In the theory of belief functions, a great number of combination rules are used to summarize a set of mass functions into only one. Let $s_1$ and $s_2$ be two distinct and cognitively independent sources providing two different mass functions $m_1$ and $m_2$ defined on the same frame of discernment $\Omega$. Combining these mass functions induces a third one $m_{12}$ defined on the same frame of discernment $\Omega$.\\
\indent There is a great number of combination rules \cite{Dubois88,Martin07b,Murphy00a,Smets94a,Yager87,Lefevre13a}, but we enumerate in this section only Dempster, conjunctive, disjunctive, Yager, Dubois and Prade, mean, cautious and bold combination rules.
The first combination rule was proposed by Dempster in \cite{Dempster67a} to combine two distinct mass functions $m_1$ and $m_2$ as follows:
\begin{equation}
\label{RDEmpster}
m _{1\oplus 2}(A)=(m_1 \oplus m_2)(A)=
\left\{
\begin{tabular}{ll}
$\frac{\displaystyle{\sum _{B\cap C=A}} m_1(B)\times m_2(C)}{1-\displaystyle{\sum _{B\cap C=\emptyset}} m_1(B)\times m_2(C)}$&$\forall A\subseteq\Omega,\hspace{0.1cm}A\neq \emptyset$\\
0&$if\hspace{0.1cm}A=\emptyset$\\
\end{tabular}
\right.
\end{equation}
The \textsc{bbm} of the empty set is null ($m(\emptyset)=0$). This rule verifies the normality condition and works under a \textit{closed world} where $\Omega$ is exhaustive.\\
\indent In order to solve the problem highlighted by Zadeh's counter example \cite{Zadeh84a} where Dempster's rule of combination produced unsatisfactory results, many combination rules appeared.
Smets \cite{Smets90a} proposed an \textit{open world} where a positive mass can be allocated to the empty set. Hence the conjunctive rule of combination for two mass functions $m_1$ and $m_2$ is defined as follows:
\begin{equation}
\label{Conjunctive}
\displaystyle{m _{1\ocap2}}(A)=(m_1\ocap m_2)(A)=\sum_{B\cap C=A} m_1(B)\times m_2(C)
\end{equation}
Even if Smets \cite{Smets92a} interpreted the \textsc{bbm}, $\displaystyle{m _{1\ocap2}}(\emptyset)$, as an amount of conflict between evidences that induced $m_1$ and $m_2$; that amount is not really a conflict because it includes a certain degree of auto-conflict due to the non-idempotence of the conjunctive combination \cite{Martin08a}.\\
\indent The conjunctive rule is used only when both sources are reliable. Smets \cite{Smets90a} proposed also to use a disjunctive combination when an unknown source is unreliable. The disjunctive rule of combination is defined for two \textsc{bba}s $m_1$ and $m_2$  as follows:
\begin{equation}
\displaystyle{m_{1\textcircled{\scriptsize{$\cup$}} 2}}(A)=(m_1 \textcircled{\scriptsize{$\cup$}} m_2 )(A)=\sum _{B\cup C =A} m_1 (B)\times m_2(C)
\end{equation}

Yager in \cite{Yager87} interpreted $m (\emptyset)$ as an amount of ignorance; consequently it is allocated to $\Omega$. Yager's rule of combination is also defined to combine two mass functions $m_1$ and $m_2$ as follows:
\begin{equation}
\label{Equation(6)}
\left\{
\begin{tabular}{ll}
$m_Y(X)=m_{1\textcircled{\scriptsize{$\cap$}}2}(X)$&$\forall X\subset\Omega,\hspace{0.1cm}X\neq\emptyset$\\
$m_Y(\Omega)=m_{1\textcircled{\scriptsize{$\cap$}}2}(\Omega)+m_{1\textcircled{\scriptsize{$\cap$}}2}(\emptyset)$\\
$m_Y(\emptyset)=0$\\
\end{tabular}
\right.
\end{equation}
Dubois and Prade's solution \cite{Dubois88} was to affect the mass resulting from the combination of conflicting focal elements to the union of these subsets:
\begin{equation}
\label{Equation(7)}
\left\{
\begin{tabular}{lll}
$m_{DP}(B)=m_{1\textcircled{\scriptsize{$\cap$}}2}(B)+\displaystyle{\sum_{A\cap X=\emptyset,\hspace{0.1cm}A\cup X=B}}m_1(X)m_2(A)$&$\forall A\subseteq\Omega,\hspace{0.1cm} A\neq\emptyset$\\
$m_{DP}(\emptyset)=0$\\
\end{tabular}
\right.
\end{equation}
Conjunctive, disjunctive and Dempster's rules are associative and commutative, but Yager and Dubois and Prade's rules are not associative, even if they are commutative. Unfortunately, all combination rules described above are not idempotent because $m\textcircled{\scriptsize{$\cap$}} m \neq m$ and $m\textcircled{\scriptsize{$\cup$}} m\neq m$.\\
\indent Mean combination rule detailed in \cite{Murphy00a}, $m_{Mean}$, of two mass functions $m_1$ and $m_2$ is the average of these ones. Therefore, for each focal element $A$ of $M$ mass functions, the combined one is defined as follows:
 \begin{equation}
 \label{mean}
 m_{Mean}(A)=\frac{1}{M}\sum^{M}_{i=1}m_i(A)
 \end{equation}
 Besides idempotence, this combination rule verifies normality condition ($m(\emptyset)=0$) if combined mass functions are normalized ($\forall i\in M,\hspace{0.1cm} m_i(\emptyset)= 0$). We note also that this combination rule is commutative but not associative. \\
\indent All combination rules described above work under a strong assumption of cognitive independence since they are used to combine mass functions induced by two distinct sources. This strong assumption is always assumed but never verified. Denoeux \cite{Denoeux08a}, proposed a family of conjunctive and disjunctive rules based on triangular norms and conorms. Cautious and bold rules are members of that family and combine mass functions for which independence assumption is not verified.
Cautious combination of two mass functions $m_1$ and $m_2$ issued from probably dependent sources is defined as follows:
\begin{equation}
\label{CautiousRule}
m_1\owedge m_2=\displaystyle{\mathop{\ocap_{A\subset\Omega}}}\hspace{0.1cm}A^{w_1(A)\wedge w_2(A)}
\end{equation}

Where $A^{w_1(A)}$ and $A^{w_2(A)}$ are simple support functions focused on $A$ with weights $w_1$ and $w_2$ issued from the canonical decomposition \cite{Smets95a} of $m_1$ and $m_2$ respectively, note also that $\wedge$ is a \textit{min operator} of simple support functions weights. The bold and cautious combination rules are commutative, associative and idempotent.\\
\indent To summarize, the choice of the combination rule is based on the dependence of sources. Combination rules like \cite{Dubois88,Martin07b,Murphy00a,Smets94a,Yager87} combine mass functions which sources are independent, whereas cautious, bold and mean rules are the most fitted to combine mass functions issued from dependent sources. \\
\indent In this paper, we propose a method to quantify sources' degrees of independence that may be used in a new mixed combination rule. In fact, we propose a statistical approach to learn sources' degrees of independence from all provided evidential information. Indeed, two sets of evidential information assessed by two different sources are classified into two sets of clusters. Clusters of both sources are matched and the independence of each couple of matched clusters is quantified in order to estimate sources' degrees of independence. Therefore, a clustering technique is used to gather similar objects into the same cluster in order to study the source's overall behavior. Before introducing our learning method, we detail in the next section the evidential clustering algorithm that will be used in the learning of sources' degrees of independence.
\section{Evidential clustering}
\label{EC}
In this paper, we propose a new clustering technique to classify objects; their attributes values are evidential and classes are unknown. Proposed clustering algorithm uses a distance on belief functions given by Jousselme et al. \cite{Jousselme01a} such as proposed by Ben Hariz et al. \cite{BenHariz06}. \\
\indent Ben Hariz et al. \cite{BenHariz06} detailed a belief $K$-modes classifier in which Jousselme distance \cite{Jousselme01a} is adapted to quantify distances between objects and clusters modes. These are sets of mass functions; each one is the combination of an attribute's values of all objects classified into that cluster. An object is attributed to the cluster having the minimum distance to its mode.\\
\indent Temporal complexity of clustering algorithm proposed by Ben Hariz et al. \cite{BenHariz06} is quite high as clusters modes and distances are computed in each iteration. The combination by the mean rule to compute modes values leads to mass functions with a high number of focal elements. Hence, the bigger the cluster is, the least significant is the distance.\\
\indent We propose a clustering technique to classify objects that attributes values are uncertain. However uncertainty is modeled with the theory of belief functions detailed in Section \ref{BFT}. In the proposed algorithm, we do not use any cluster mode to avoid the growth of focal elements number in clusters modes. Temporal complexity is also significantly reduced because all distances are computed only once.\\
\indent In this section, $K$ is the number of clusters $Cl_k$ ($1\leq k\leq K$); $n$ is the number of objects to be classified;
$n_k$ is the number of objects classified into cluster $Cl_k$;
$o_i$ are objects to classify $o_i: 1\leq i\leq n$; $c$ is the number of evidential attributes $a_j:\hspace{0.1cm}1\leq j\leq c$ which domains are $\Omega_{a_j}$ and finally $m_{ij}$ is a mass function value of attribute ``$j$'' for object ``$i$''. Mass functions $m_{ij}$ can be certain, probabilistic, possibilistic, evidential and even missing.

To classify objects $o_i$ into $K$ clusters, we use a clustering algorithm with a distance on belief functions given by \cite{Jousselme01a}. The number of clusters $K$ is assumed to be known. Proposed clustering technique is based on a distance which quantifies how much is far an object $o_i$  from a cluster $Cl_k$. This distance is the mean of distances between $o_i$, and all objects $o_q$ that are classified into cluster $Cl_k$ as follows:
\begin{equation}
\label{dist}
D(o_i,Cl_k)=\frac{1}{n_k}\sum_{q=1}^{n_k}dist(o_i,o_q)
\end{equation}
and 
\begin{equation}
dist(o_i,o_q)=\frac{1}{c}\sum_{j=1}^{c}d(m_{ij},m_{qj})
\end{equation}
with~:
\begin{equation}
\label{JDist}
d(m_{ij},m_{qj})=\sqrt{\frac{1}{2}(m_{ij}-m_{qj})^t\underline{\underline{D}}(m_{ij}-m_{qj})}
\end{equation}
such that~:
\begin{equation}
\underline{\underline{D}}(A,B)=\left\{
\begin{tabular}{ll}
$1$& if $A=B=\emptyset$\\
$\frac{\mid A\cap B\mid}{\mid A\cup B\mid}$&$ \forall A,B \in 2^{\Omega_{a_j}}$\\
\end{tabular}
\right.
\end{equation}

\begin{figure}[ht]
\begin{center}
\includegraphics[scale=0.45,trim = 8cm 1cm 8cm 1cm]{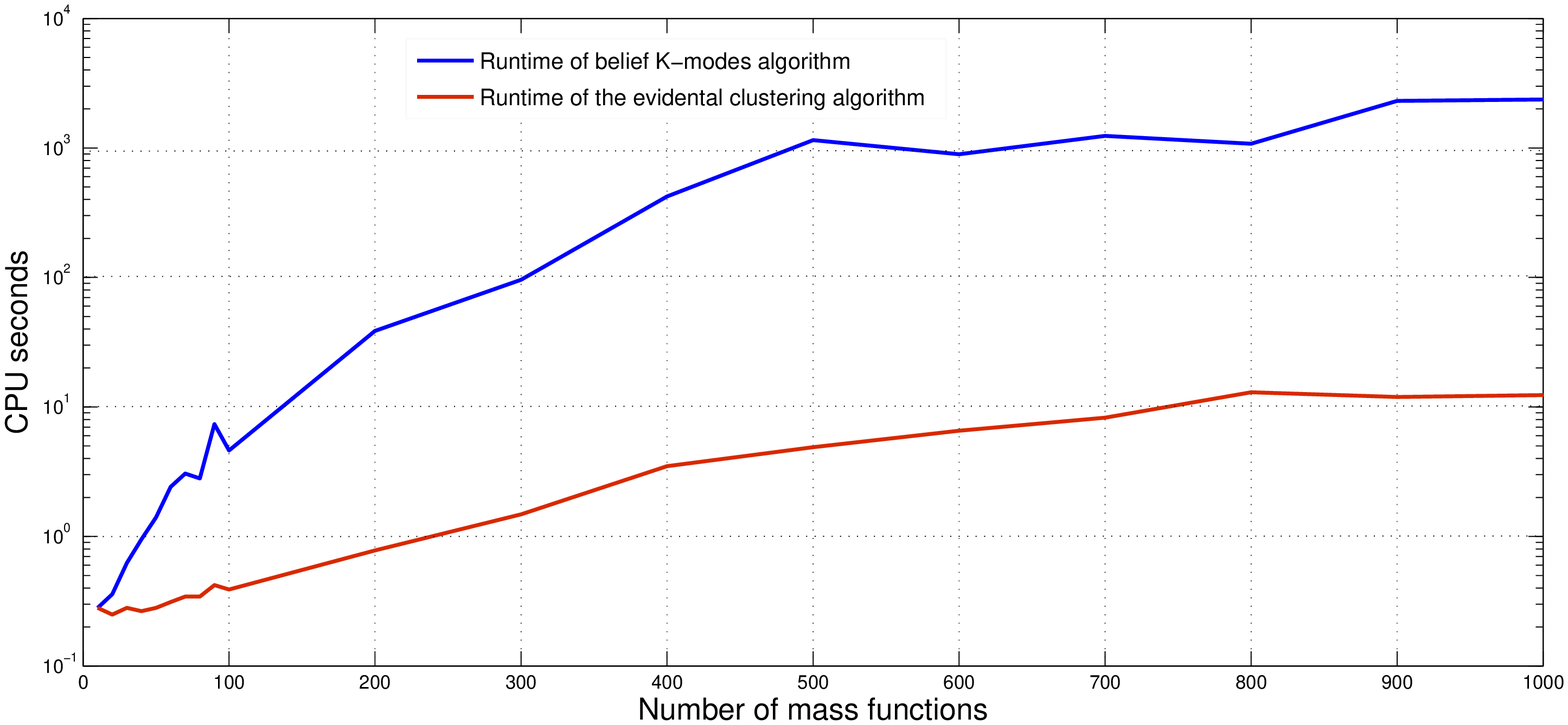}
\caption{Run-time optimization of the evidential clustering and the belief $K$-modes \cite{BenHariz06} according to $n\in[10,1000]$, $\mid\Omega_{a_j}\mid=5$ and $K=5$}
\label{Tcomp4}
\end{center}
\end{figure}

Each object is affected to the most similar cluster in an iterative way till reaching an unchanged cluster partition. It is obvious that clusters number $K$ must be known.
Temporal complexity of the proposed algorithm is significantly optimized as pairwise distances are computed once a time from the beginning. We do not use any cluster mode. Consequently, there will be no problem of increasing number of focal elements because attributes values are not combined.
Indeed, the evidential clustering algorithm provides a cluster partition that minimizes distances between objects into the same cluster and maximizes the distance between objects classified into different clusters. The main asset of the evidential clustering algorithm according to the belief $K$-modes proposed by Ben Hariz et al. \cite{BenHariz06} is the optimization of the temporal complexity. In fact, run-time of the evidential clustering algorithm is improved. The optimization of run-time depends on the size of the frame of discernment $|\Omega_{a_j}|$, the number of clusters $K$ and number of objects $n$. For example, figure \ref{Tcomp4} shows a big gain in the run-time of evidential clustering according to the belief $K$-modes when the number of mass functions varies, $n\in[10,1000]$.
Temporal complexity of the evidential clustering algorithm is optimized and that optimization is especially noticed when the number of mass functions to classify is high and also when the frame of discernment contains many hypotheses. Thanks to the improve of the temporal complexity, this clustering algorithm is used in the following sections.
\section{Learning sources independence degree}
In this section we extend paper \cite{Chebbah12a} for many sources, and propose a combination rule emphasizing sources independence degree.
In the theory of probabilities, two hypotheses $X$ and $Y$ are assumed to be statistically independent if $P(X\cap Y)=P(X)\times P(Y)$ or $P(X | Y)=P(X)$. In the context of the theory of belief functions, Shafer \cite{Shafer76} defined cognitive and evidential independence. 
\vspace{-0,3cm}
\paragraph{\bfseries{Definition 1}} ``Two frames of discernment may be called cognitively independent with respect to the evidence if new evidence that bears on only one of them will not change the degree of support for propositions discerned by the other'' \footnote{\cite{Shafer76}, page 149}.
\vspace{-0,05cm}\\
\indent The cognitive independence is a \textit{weak independence}; two variables are independent with respect to a mass function if new evidence that bears on only one of the two variables does not change propositions discerned by the other one. For two variables $X$ and $Y$ such that $\Omega_X$ and $\Omega_Y$ their domains (frames of discernment) and $\Omega_X\times \Omega_Y$ the product space of domains $\Omega_X$ and $\Omega_Y$. Variables $X$ and $Y$ are cognitively independent with respect to $m^{\Omega_X\times \Omega_Y}$ if:
\begin{equation}
pl^{\Omega_X\times \Omega_Y}(x,y)=pl^{\Omega_X\times \Omega_Y\downarrow \Omega_X}(x)\times pl^{\Omega_X\times \Omega_Y\downarrow \Omega_Y}(y)
\end{equation}
Note that $\Omega_X\times \Omega_Y\downarrow \Omega_X$ is the marginalization of $\Omega_X\times \Omega_Y$ in $\Omega_X$ \cite{Smets94a,Smets97c}.\\
Shafer \cite{Shafer76} defined also a \textit{strong independence} called evidential independence as follows:
\vspace{-0,2cm}
\paragraph{\bfseries{Definition 2}} ``Two frames of discernment are evidentially independent with respect to a support function if that support function could be obtained by combining evidence that bears on only one of them with evidence that bears on only the other''.
\vspace{-0,4cm}\\
\indent Two variables are evidentially independent if their joint mass function can be obtained by combining marginal mass functions that bears on each one of them. Variables $X$ and $Y$ are evidentially independent with respect to $m^{\Omega_X\times \Omega_Y}$ if:
\begin{equation}
\left\{
\begin{tabular}{l}
$pl^{\Omega_X\times \Omega_Y}(x,y)=pl^{\Omega_X\times \Omega_Y\downarrow \Omega_X}(x)\times pl^{\Omega_X\times \Omega_Y\downarrow \Omega_Y}(y)$\\
$bel^{\Omega_X\times \Omega_Y}(x,y)=bel^{\Omega_X\times \Omega_Y\downarrow \Omega_X}(x)\times bel^{\Omega_X\times \Omega_Y\downarrow \Omega_Y}(y)$
\end{tabular}
\right.
\end{equation}

Independence can also be defined in terms of \textit{irrelevance}. The knowledge of the value of one variable does not change the belief on the other one. In the theory of belief functions, irrelevance is based on the conditioning. Variables $X$ and $Y$ are irrelevant with respect to $m$, $IR_m(X,Y)$ if the marginal mass function on $X$ is obtained by conditioning the joint mass function on values $y$ of $Y$ and marginalizing this conditioned joint mass function on $X$:
\begin{equation}
{m^{\Omega_X\times \Omega_Y\downarrow \Omega_X}_{[y]}}(x)\propto m^{\Omega_X\times \Omega_Y\downarrow \Omega_X}(x)
\end{equation}
Note that proportionality $\propto$ is replaced by equality when ${m^{\Omega_X\times \Omega_Y\downarrow \Omega_X}_{[y]}}$ and $m^{\Omega_X\times \Omega_Y\downarrow \Omega_X}$ are normalized.

Doxastic independence is especially proposed in the theory of belief functions by \cite{BenYaghlane02a,BenYaghlane02b} and it is defined as follows:
\vspace{-0,3cm}
\paragraph{\bfseries{Definition 3}} ``Two variables are considered as doxastically independent only when they are irrelevant and this irrelevance is preserved under Dempster's rules of combination''.\\
\vspace{-0,8cm}\\
\indent In other words, two variables $X$ and $Y$ are doxastically independent if they are irrelevant with respect to $m\oplus m_0$ when they are irrelevant with respect to $m$ and $m_0$. Indeed, if $X$ and $Y$ are irrelevant according to any mass function $m$ and if they are also irrelevant with respect to another mass function $m_0$; they are assumed to be doxastically independent if they are irrelevant with respect to the orthogonal sum of $m$ and $m_0$ . Thus, if $IR_m(X,Y)$, $IR_{m_0}(X,Y)$ and $IR_{m\oplus m_0}(X,Y)$ is verified then $X$ and $Y$ are doxastically independent.

This paper is not focused on variables independence \cite{BenYaghlane02a, BenYaghlane02b, Shafer76} but on \textit{sources independence}. Sources independence is computed according to a set of different belief functions provided by each source separately. Sources are dependent when all their beliefs are correlated, there is a link between all mass functions they provide. This problem is not tackled till now, we noticed a lack of references treating this problem.
To study sources independence, a great number of mass functions provided by both sources is needed. This set of mass functions must be defined on the same frame of discernment according to the same problems. For example, two distinct doctors provide $n$ diagnoses in the examination of the same $n$ patients. In that case, the frame of discernment contains all diseases and is already the same for both doctors. We define sources independence as follows:
\vspace{-0,2cm}
\paragraph{\bfseries{Definition 4}}{Two sources are cognitively independent if they do not communicate and if their evidential corpora are different.}
\vspace{-0,2cm}
\paragraph{\bfseries{Definition 5}}{Evidential corpus is the set of all pieces of evidence held by a source.}
\vspace{-0,2cm}\\
\indent Not only communicating sources are considered dependent but also sources having the same background of knowledge since their beliefs are correlated. The aim of estimating sources independence is either to guide the choice of combination rules when aggregating their beliefs, or to integrate this degree of independence in a new combination rule.

In this paper, mass functions provided by two sources are studied in order to reveal any dependence between them.
In the following, we define an independence measure  $\Id$, ($\Id(s_1,s_2)$), as the independence of $s_1$ on $s_2$ verifying the following axioms:
\begin{enumerate}
\item{Non-negativity:} The independence of a source $s_1$ on another source $s_2$, $\Id(s_1,s_2)$ cannot be negative, it is either positive or null.
\item{Normalization:} The degree of independence $\Id$ is a degree over $[0,1]$, it is null when the first source is dependent on the second one, equal to $1$ when it is completely independent and a degree from $[0,1]$ otherwise.
\item Non-symmetry: In the case where $s_1$ is independent on $s_2$, $s_2$ is not necessarily independent on $s_1$. Even if $s_1$ and $s_2$ are mutually independent, degrees of independence are not necessarily equal.
\item Identity: Any source is completely dependent on itself and $\Id(s_1,s_1)=0$.
\end{enumerate}
 If $s_1$ and $s_2$ are independent, there will be no correlation between their mass functions. 
 The main idea of this paper is: First, classify mass functions provided by each source separately. Then, study similarities between cluster partitions to reveal any dependence between sources. By using clustering algorithm, sources overall behavior is studied. 
 The proposed method is in three steps: First, mass functions of each source are classified. Then, similar clusters are matched. Finally, weights of linked clusters and sources independence are quantified.
\subsection{Clustering}
\label{clustering}
Clustering algorithm detailed in Section \ref{EC} is used to classify two sets of $n$ mass functions respectively provided by sources $s_1$ and $s_2$. Clustering algorithm is performed on all mass functions of $s_1$ independently of the clustering performed on those of $s_2$. We remind that all mass functions of both sources are defined on the same frame of discernment and so considered as values of only one attribute when classifying their corresponding objects. For the same example of doctors, patients are objects to classify according to an attribute \textit{disease}. Values of this attribute are mass functions defined on the frame of discernment enumerating all possible diseases.
Distance (\ref{dist}) can be simplified as follows because we have only one attribute:
\begin{equation}
\label{distS}
D(o_i,Cl_k)=\frac{1}{n_k}\sum_{q=1}^{n_k}d(m_i,m_q)
\end{equation}
In this paper, we fix the number of clusters to the number of hypotheses in the frame of discernment. 
In a classification point of view, number of hypotheses is the number of possible classes. For example, the frame of discernment of the attribute disease enumerates all possible diseases. Hence, when a  doctor examines a patient, he gives a mass function as a classification of the patient in some possible diseases. 
\subsection{Cluster matching}
After clustering technique, both mass functions provided by $s_1$ and $s_2$  are distributed separately on $K$ clusters. 
In this section, we try to find a mapping between clusters in order to link those containing the same objects. If clusters are perfectly linked, meaning all objects are classified similarly for both sources, we can conclude that sources are dependent as they are choosing similar focal elements (not contradictory at least) when providing mass functions for same objects. If clusters are weakly linked, sources choose similar focal elements for different objects and so they are independent. Clusters independence degree is proportional to the number of objects similarly classified. More clusters contain the same objects, more they are dependent as they are correlated.

We note $Cl_{k_1}^1$ where $1\leq k_1 \leq K$ for clusters of $s_1$ and $Cl_{k_2}^2$ where $1\leq k_2 \leq K$ for those of $s_2$. The similarity between two clusters $Cl_{k_1}^1$ and $Cl_{k_2}^2$ is the proportion of objects simultaneously classified into $Cl_{k_1}^1$ and $Cl_{k_2}^2$:
\begin{equation}
\beta^i_{k_ik_j}=\beta^i(Cl_{k_i}^i,Cl_{k_j}^j)=\frac{\mid Cl^{i}_{k_i} \cap Cl^{j}_{k_j} \mid}{\mid Cl^{i}_{k_i}\mid}
\end{equation}

with ${i,j}\in\{1,2\}$ and $i\neq j$. $\beta^1_{k_1k_2}$ quantifies a proportion of objects classified simultaneously in clusters $Cl^1_{k_1}$ and $Cl^2_{k_2}$ with regard to objects in $Cl^1_{k_1}$, analogically $\beta^2_{k_2k_1}$ is a proportion of objects simultaneously in $Cl^1_{k_1}$ and $Cl^2_{k_2}$ with regard to those in $Cl^2_{k_2}$. Note that $\beta^1_{k_1k_2}\neq \beta^2_{k_2k_1}$ since the number of objects classified into  $Cl^1_{k_1}$ and $Cl^2_{k_2}$ are different ($\mid Cl^1_{k_1}\mid \neq \mid Cl^2_{k_2}\mid$).

We remind that $\beta^1$ are similarities towards $s_1$ and $\beta^2$ are those towards $s_2$. It is obvious that $\beta^i(Cl^i_{k_i},Cl^j_{k_j})=0$ when $Cl^i_ {k_i}$ and $Cl^j_{k_j}$ do not contain any common object; however they are completely different. $\beta^i(Cl^i_{k_i},Cl^j_{k_j})=1$ when these clusters are strongly similar so they contain the same objects. 
A similarity matrix $M_1$ containing similarities of clusters of $s_1$ according to those of $s_2$ ($\beta^1$), and $M_2$ the similarity matrix between clusters of $s_2$ and those of $s_1$ ($\beta^2$) are defined as follows:
\begin{eqnarray}
M_1= \begin{pmatrix}
\beta^1_{11}&\beta^1_{12}&\ldots& \beta^1_{1K}\\ 
\ldots&\ldots&\ldots&\ldots\\
\beta^1_{k1}&\beta^1_{k2}&\ldots& \beta^1_{kK}\\   
\ldots&\ldots&\ldots&\ldots\\
\beta^1_{K1}&\beta^1_{K2}&\ldots& \beta^1_{KK}\\  
\end{pmatrix}
& \quad \mbox{and} \quad
M_2= \begin{pmatrix}
\beta^2_{11}&\beta^2_{12}&\ldots& \beta^2_{1K}\\ 
\ldots&\ldots&\ldots&\ldots\\
\beta^2_{k1}&\beta^2_{k2}&\ldots& \beta^2_{kK}\\   
\ldots&\ldots&\ldots&\ldots\\
\beta^2_{K1}&\beta^2_{K2}&\ldots& \beta^2_{KK}\\  
\end{pmatrix}
\end{eqnarray}

We note that $M_1$ and $M_2$ are different since $\beta^1_{k_1k_2}\neq \beta^2_{k_2k_1}$. Clusters of $s_1$ are matched to those of $s_2$ according to maximum of $\beta^1$ such that each cluster $Cl^1_{k_1}$ is linked to only one cluster $Cl^2_{k_2}$ and each cluster $Cl^2_{k_2}$ has only one cluster $Cl^1_{k_1}$ linked to it. The idea is to link iteratively clusters having the maximal $\beta^1$ in $M_1$ then eliminate these clusters and the corresponding line and column from the matrix until having a bijective cluster matching. Algorithm \ref{cluster matching} details cluster matching process.
We note that different matchings are obtained for $s_1$ and $s_2$ because $M_1$ and $M_2$ are different. 
\begin{algorithm}[h]
\caption{Cluster matching}
\label{cluster matching}
\begin{algorithmic}[1]
\REQUIRE Similarity matrix $M$ .
\WHILE {$M$ is not empty}
   \STATE Find $max(M)$ and indexes $c$ and $l$ of clusters having this maximal similarity.
	 \STATE Map clusters $l$ and $c$.
	 \STATE Delete line $l$ and column $c$ from $M$.
\ENDWHILE
\RETURN Cluster matching.
\end{algorithmic}
\end{algorithm}
This algorithm is iterative and the number of iteration is equal to the number of clusters $K$. Even if this algorithm is quite simple, it provides a matching of clusters in order to compare evidential information provided by both sources. The assignment algorithm proposed in \cite{Munkers57}  for square matrices and that for rectangular matrices \cite{Bourgeois71} can also be used to minimize the dissimilarity between matched clusters. Other methods for cluster matching \cite{Wemmert02} and \cite{Gancarski05} can also be used.
\subsection{Cluster independence}
\begin{figure}[ht]
\begin{center}
\includegraphics[scale=0.48,trim = 9cm 3cm 9cm 1cm]{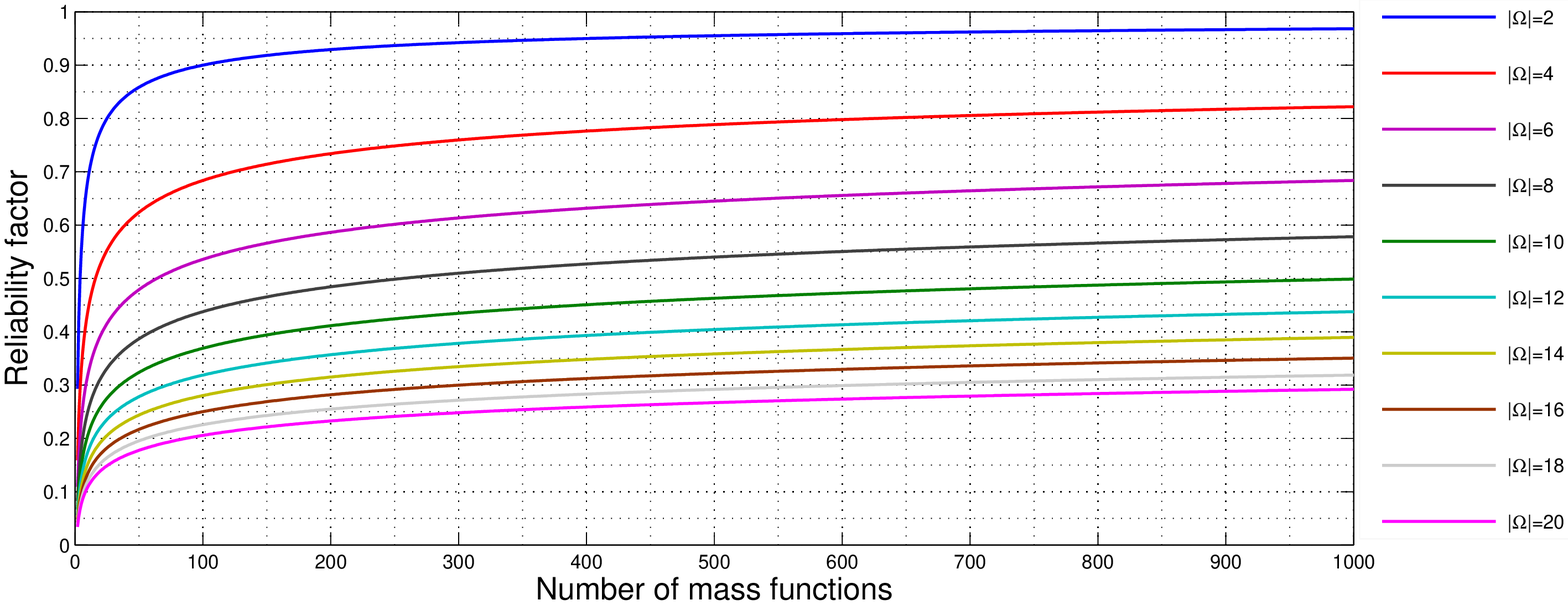}\caption{Reliability factors $\alpha^i_{k_i}$}\label{fact}\end{center}\end{figure}

Once cluster matching is obtained, a degree of independence/dependence of matched clusters is quantified in this step.
 A set of matched clusters is obtained for both sources and a mass function can be used to quantify each couple of clusters independence. Assume that cluster $Cl^1_{k_1}$ is matched to $Cl^2_{k_2}$, a mass function $m^{\Omega_I}$\footnote{We note the frame of discernment in the mass functions to avoid confusion.} defined on the frame of discernment $\Omega_I=\{Dependent\hspace{0.2cm}Dep, Independent\hspace{0.2cm}Ind\}$ describes how much this couple of clusters is independent or dependent as follows:
\begin{equation}
\label{CD}
\left\{
\begin{tabular}{ll}
$m^{\Omega_I,i}_{{k_i}{k_j}}(Dep)=\alpha^i_{k_i}  \beta^1_{{k_i}{k_j}}$\\
$m^{\Omega_I,i}_{{k_i}{k_j}}(Ind)=\alpha^i_{k_i}  (1-\beta^1_{{k_i}{k_j}})$\\
$m^{\Omega_I,i}_{{k_i}{k_j}}(Dep\cup Ind)=1-\alpha^i_{k_i}$\\
\end{tabular}
\right.
\end{equation}
A mass function quantifies the degree of independence of each couple of clusters according to each source; $m^{\Omega_I,i}_{{k_i}{k_j}}$ is a mass function for the independence of each linked clusters $Cl^i_{k_i}$ and $Cl^j_{k_j}$ according to $s_i$ with $i,j\in\{1,2\}$ and $i\neq j$.
Coefficient $\alpha^{i}_{k_i}$ is used to take into account of number of mass functions in each cluster $Cl_{k_i}$ of the source $i$. Reliability factor $\alpha^{i}_{k_i}$ is not the reliability of any source but it can be seen as the reliability of the clusters independence estimation. Consequently, independence estimation is more reliable when clusters contain enough mass functions. For example, assume two clusters; one containing only one mass function and the second one containing $100$ mass functions. It is obvious that the independence estimation of the second cluster is more precise and significant than the independence estimation of the first one. \\
Reliability factors $\alpha^{i}_{k_i}$ are proportional to the number of hypotheses in the frame of discernment $\mid \Omega\mid$, and the number of objects classified in $Cl^{i}_{k_i}$ as follows:
\begin{equation}
\label{DF}
\alpha^i_{k_i}=f(\mid\Omega\mid, \mid Cl^i_{k_i}\mid)
\end{equation}
The bigger $\mid \Omega\mid$ is, the more mass functions are needed to have a reliable cluster independence estimation. For example, if $\mid\Omega \mid=5$ then there are $2^5$ possible focal elements, also independence estimation of a cluster containing $20$ objects cannot be precise. No existing method to define such function $f$. Hence, we use simple heuristics as follows:
\begin{equation}
\alpha^i_{k_i}=1-\frac{1}{\mid Cl^i_{k_i}\mid^\frac{1}{\mid \Omega\mid}}
\end{equation}
As shown in figure \ref{fact}, if $\mid\Omega\mid$ and number of mass functions in a cluster are big enough, cluster independence mass function is almost not discounted. Reliability factor is an increasing function of $\mid\Omega\mid$ and $\mid Cl^i_{k_i}\mid$ which favors big clusters\footnote{Big clusters are those containing enough mass functions according to $\mid\Omega\mid$.}.

\subsection{Sources independence}
Obtained mass functions quantify each matched clusters independence according to each source. Therefore, $K$ mass functions are obtained for each source such that each mass function quantifies the independence of each couple of matched clusters. The combination of $K$ mass functions for each source using the mean, defined by equation (\ref{mean}), is a mass function $m^{\Omega_I}$ defining the whole independence of one source on another one: 
\begin{equation}
\begin{tabular}{ll}
 $m^{\Omega_I,s_i}(A)=\frac{1}{K}\displaystyle{\sum^{K}_{k_i=1}}m^{\Omega_I,i}_{{k_i}{k_j}}(A)$&$\forall A\subseteq 2^\Omega$\\
\end{tabular}
 \end{equation}

With $k_j$ is the cluster matched to $k_i$ according to $s_i$. Two different mass functions $m^{\Omega_I,s_1}$ and $m^{\Omega_I,s_2}$ are obtained for $s_1$ and $s_2$ respectively. We note that $m^{\Omega_I,s_1}$ is the combination of $K$ mass functions representing the independence of matched clusters according to $s_1$ defined using equation (\ref{CD}). Mass functions $m^{\Omega_I,s_1}$ and $m^{\Omega_I,s_2}$ are different since cluster matchings are different which verifies the axiom of non-symmetry. $\beta^1_{{k_1}{k_2}},\beta^2_{{k_2}{k_1}}\in[0,1]$ verify the non-negativity and the normalization axioms. 
Finally, pignistic probabilities are computed from these mass functions in order to decide about sources independence $\Id$ such that:
\begin{equation}
\label{ID}
\left\{
\begin{tabular}{ll}
$\Id(s_1,s_2)=\BetP(Ind)$\\
 $\overline{\Id}(s_1,s_2)=\BetP(Dep)$
\end{tabular}
\right.
\end{equation} 
If $\Id(s_1,s_2)>\overline{\Id}(s_1,s_2)$ we claim that sources $s_1$ and $s_2$ are independent otherwise they are dependent.
\subsection{General case}
The method detailed above estimates the independence of one source on another one. Independence measure is non-symmetric because if a source $s_1$ is independent on a source $s_2$ then $s_2$ is not necessarily independent on $s_1$ and even if it is the case, degrees of independence are not necessarily the same.\\
\indent It is wise to choose the minimum independence from $\Id(s_1,s_2)$ and $\Id(s_2,s_1)$ as the overall independence. Consequently, if at least one of two sources is dependent on the other, then sources are considered dependent. In other words, two sources are independent only if they are mutually independent.
Hence, overall independence that is denoted $I(s_1,s_2)$ is given by:
\begin{equation}
I(s_1,s_2)=min(\Id(s_1,s_2),\Id(s_2,s_1))
\end{equation}
We note that $I(s_1,s_2)$ is non-negative, normalized, symmetric and identical.\\
We define an independence measure, noted $I$, generalizing the independence for more than two sources verifying the following axioms:
\begin{enumerate}
\item{Non-negativity:~}Many sources independence $\{s_1,s_2,s_3, \ldots,s_{ns}\}$, noted \break $\Id(s_1,s_2, \ldots, s_{ns})$ cannot be negative, it is either positive or null.
\item{Normalization:~}Sources independence $I$ is a degree in $[0,1]$. The minimum $0$ is reached when sources are completely dependent and the maximum $1$ is reached when they are completely independent.
\item {Symmetry:~}$I(s_1, s_2, s_3, \ldots, s_{ns})$ is the sources' overall independence and\break $I(s_1, s_2, s_3, \ldots, s_{ns})=I(s_2,s_1,s_3,\ldots,s_{ns})=I(s_3,s_1,s_2,\ldots,s_{ns})$.
\item {Identity:~}$I(s_1,s_1,s_1)=0$. It is obvious that any source is completely dependent on itself.
\item {Increasing with inclusion:~}$I(s_1,s_2)\leq I(s_1,s_2,s_3)$, more there are sources, more they are likely to be independent.
\end{enumerate}

To compute the overall independence of $ns$ sources $\{s_1,s_2,\ldots, s_{ns}\}$, independencies of pairs of sources are computed and the maximum\footnote{The maximum is used to insure the property of increasing with inclusion.} independence is the sources overall independence:
\begin{equation}
\label{indmanysources}
I(s_1,s_2,\ldots,s_{ns})=max(I(s_i,s_j)),\hspace{1cm}\forall i\in[1,ns]\hspace{0.2cm},j\in]i,ns]
\end{equation}
or equivalently:
\begin{equation}
I(s_1,s_2,\ldots,s_{ns})=max(min(I_d(s_i,s_j),I_d(s_j,s_i))),\hspace{1cm}\forall i,j\in[1,ns]\hspace{0.2cm}i\neq j
\end{equation}
Independence degree of sources is then integrated in the combination step using the following mixed combination rule.
\section{Combination rule}
Combination rules using conjunctive and/or disjunctive rules such as \cite{Dubois88,Martin07b,Murphy00a,Smets94a,Yager87} are used when sources are completely independent but cautious and bold rules \cite{Denoeux08a} tolerate redundant information and consequently can be used to combine mass functions which sources are dependent. In the combination step, sources dependence or independence hypothesis is intuitively made without any possibility of check. Sources independence degree is neither $0$ nor $1$ but a level over $[0,1]$. The main question is \enquote{which combination rule to use when combining partially independent$\backslash$dependent mass functions?}\\
\indent In this paper, we propose a new mixed combination rule using conjunctive and cautious rules detailed in equations \eqref{Conjunctive} and \eqref{CautiousRule}. In the case of totally dependent sources (where independence is $0$), the cautious and proposed mixed combination rules are similar; whereas in the case of totally independent sources (independence is $1$), the conjunctive and proposed combination rules are similar. In the case of an independence degree in $]0,1[$, combined mass function is the average of conjunctive and cautious combinations weighted by sources' independence degree.\\   
\indent Assume that two sources $s_1$ and $s_2$ are independent with a degree $\gamma$ such that $\gamma=I(s_1,s_2)$; $m_1$ and $m_2$ are mass functions provided by $s_1$ and $s_2$. The proposed mixed combination rule is defined as follows:
\begin{equation}
\begin{tabular}{ll}
$m_{Mixed}(A)=\gamma*m_{\ocap}(A)+(1-\gamma)*m_{\owedge}(A)$,& $\forall A\subseteq\Omega$
\end{tabular}
\end{equation}
The degree of independence of a set of sources is given by equation \eqref{indmanysources}, and the mixed combination of a set of mass functions $\{m_1,m_2,\ldots,m_{ns}\}$ provided by sources $\{s_1,s_2,\ldots,s_{ns}\}$ is also a weighted average such that:
\begin{equation}
\gamma=I(s_1,s_2,\ldots,s_{ns})
\end{equation}
\textbf{Properties of the proposed mixed combination rule:}
\begin{itemize}
\item{Commutativity:~}Conjunctive and cautious rules are commutative. Independence measure is symmetric because sources' degree of independence is the same for a set of sources. Then the proposed rule is commutative.
\item{Associativity:~}Conjunctive and cautious rule are associative but the proposed rule is not because independence degree of $n$ sources and $n+1$ ones is not necessarily the same.
\item{Idempotent:~}Degree of independence of one source to itself is $0$, in that case the proposed rule is equivalent to the cautious rule. As the cautious rule is idempotent, it is the case of the proposed mixed rule. 
\item{Neutral element:~}Mixed combination rule does not have any neutral element.
\item{Absorbing element:~}No absorbing element also.
\end{itemize}

\paragraph{Example} Assume a frame of discernment $\Omega=\{a, b, c\}$ and two sources $s_1$ and $s_2$ providing two mass functions $m_1$ and $m_2$. Table \ref{tabcomb} illustrates conjunctive and cautious combinations as well as mixed combination in the cases where $\gamma=0$, $\gamma=0.3$, $\gamma=0.6$ and $\gamma=1$. When $\gamma=0$, mixed and cautious combinations are equivalent; when $\gamma=1$, mixed and conjunctive combinations are equivalent, otherwise it is a weighted average by $\gamma\in]0,1[$.

\begin{figure}[ht]
\begin{center}
\includegraphics[scale=0.35,trim = 15cm 2cm 15cm 1cm]{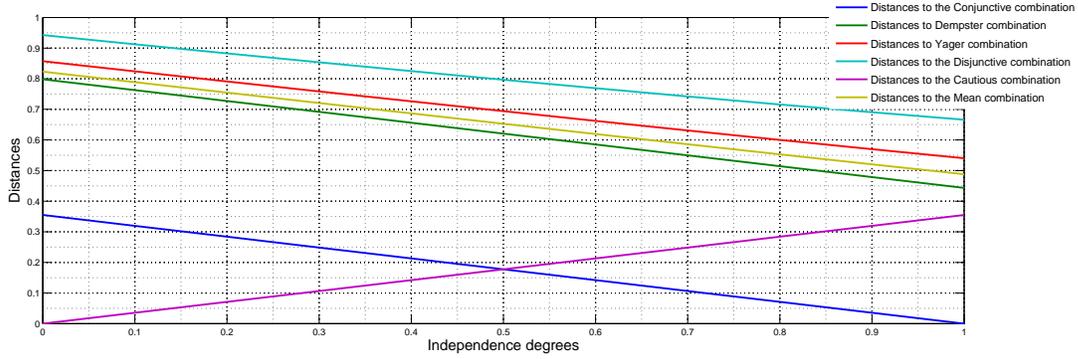}
\caption{Distances between combined mass functions}
\label{Fig2}
\end{center}
\end{figure}

Finally, to illustrate the proposed mixed combination rule and compare it to other combination rules, three mass functions are generated randomly using algorithm \ref{Mass generating}. These mass functions are combined with conjunctive, Dempster, Yager, disjunctive, cautious and mean combination rules. They are also combined with the mixed combination rule with different independence levels.\\
Figure \ref{Fig2} illustrates distances\footnote{Jousselme distance detailed in equation \eqref{JDist}.} between the mixed combination with several degrees of independence and combined mass functions using conjunctive, Dempster, Yager, disjunctive, cautious and mean combination rules. Distances between mixed combination with several independence degrees; and Yager, disjunctive, mean and Dempster's rules are linear and decreasing proportionally to $\gamma$. 
\begin{table}[ht]
\caption{Combination of two mass functions}
\label{tabcomb}
\begin{center}
\begin{tabular}{| l | l | l | >{\columncolor{lgray}}l |>{\columncolor{light-gary}} l | >{\columncolor{lgray}}l | l | l |>{\columncolor{light-gary}} l |}
\hline
$2^\Omega$&$m_1$&$m_2$&$m_{\owedge}$&$m_{\ocap}$&$m_{Mixed}$&$m_{Mixed}$&$m_{Mixed}$&$m_{Mixed}$\\
&&&&&$\gamma=0$&$\gamma=0.3$&$\gamma=0.6$&$\gamma=1$\\
\hline
$\emptyset$&0&0&0.1071&0.06&0.1071&0.093&0.0789&0.06\\
$a$&0.3&0.3&0.2679&0.45&0.2679&0.3225&0.3771&0.45\\
$b$&0&0&0&0&0&0&0&0\\
$a\cup b$&0&0&0&0&0&0&0&0\\
$c$&0.2&0&0.1786&0.14&0.1786&0.167&0.1554&0.14\\
$a\cup c$&0.2&0.4&0.2551&0.26&0.2551&0.2566&0.2580&0.26\\
$b\cup c$&0&0&0&0&0&0&0&0\\
$a\cup b \cup c$&0.3&0.3&0.1913&0.09&0.1913&0.1609&0.1305&0.09\\
\hline
\end{tabular}
\end{center}
\end{table}

\section{Experiments}
Because of the lack of real evidential data, we use generated mass functions to test the method detailed above. Moreover, it is difficult to simulate all situations with all possible combinations of focal elements for several degrees of independence between sources. First, we generate two sets of mass functions for two sources $s_1$ and $s_2$; then we illustrate for three sources.  
\subsection{Generated data depiction}
Generating sets of $n$ mass functions for several sources depends on sources independence. We discern cases of independent and dependent sources.
\subsubsection{Independent sources}
In general, to generate mass functions some information are needed: the number of hypotheses in the frame of discernment, $\mid\Omega\mid$ and the number of mass functions. We note that number of focal elements, and masses are chosen randomly. 

In the case of independent sources, masses can be anywhere and focal elements of both sources are chosen independently. Mass functions of $s_1$ and $s_2$ are generated following algorithm \ref{Mass generating}. We note that focal elements, their number and \textsc{bbm}s are chosen randomly according to the universal low.

\begin{algorithm}[ht]
\caption{Independent mass functions generating}
\label{Mass generating}
\begin{algorithmic}[1]
\REQUIRE $|\Omega|$, $n:$ number of mass functions
\FOR {$i=1$ to $n$}
   \STATE Choose randomly $\mid F\mid$, the number of focal elements on $[1,| 2^\Omega |]$. 
   \STATE Choose randomly $\mid F\mid$ focal elements noted $F$.
   \STATE Divvy the interval $[0,1]$ into $|F|$ continuous sub-intervals.
   \STATE Focal elements \textsc{bbm}s are intervals sizes.
\ENDFOR
\RETURN $n$ mass functions
\end{algorithmic}
\end{algorithm}
\subsubsection{Dependent sources}
\begin{figure}[ht]
\begin{center}
\includegraphics[scale=0.7,trim = 0cm 0.5cm 0cm 0cm]{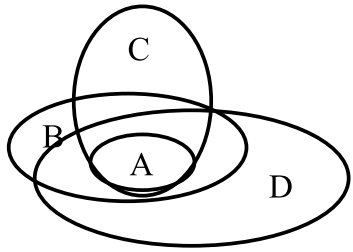}
\caption{Consistent belief function}
\label{figconsistent}
\end{center}
\end{figure}

The case of dependent sources is a bit difficult to simulate as several scenarios can occur. In this section, we will try to illustrate the most common situations.\\
Generated mass functions for dependent sources are supposed to be consistent and do not enclose any internal conflict \cite{Milan10}. Consistent mass functions contain at least one focal element common to all focal sets. Figure \ref{figconsistent} illustrates a consistent mass function where all focal elements $\{A,\hspace{0.1cm}B,\hspace{0.1cm}C,\hspace{0.1cm}D\}$ intersect.

Algorithm \ref{Consistent mass generating} generates a set of $n$ consistent mass functions\footnote{Conflict within such mass functions is null.} defined on a frame of discernment of size $\mid\Omega\mid$.  In the case of dependent sources, they are almost consistent and at least one of them is dependent on the other. To simulate the case where one source is dependent on another one, consistent mass functions of the first one are generated following algorithm \ref{Consistent mass generating}, then those of the second source are generated knowing decisions of the first one. Algorithm \ref{Dependent} generates a set of mass functions that are dependent on another set of mass functions. Dependence is due to the knowledge of other source's decisions.
\begin{algorithm}[ht]
\caption{Consistent mass functions generating}
\label{Consistent mass generating}
\begin{algorithmic}[1]
\REQUIRE $|\Omega|$, $n:$ number of mass functions
\FOR {$i=1$ to $n$}
	\STATE Choose randomly a focal set $\omega_i$ (it can be a single point) from $\Omega$.
	\STATE Find the set $S$ of all focal sets including $\omega_i$.
	\STATE Choose randomly $\mid F\mid$, the number of focal elements on $[1,| S |]$. 
	\STATE Choose randomly $\mid F\mid$ focal elements from $S$ noted $F$.
	\STATE Divvy the interval $[0,1]$ into $|F|$ continuous sub-intervals.
	\STATE \textsc{bbm}s of focal elements are intervals sizes.
\ENDFOR
\RETURN $n$ consistent mass functions
\end{algorithmic}
\end{algorithm}

\begin{algorithm}[ht]
\caption{Dependent mass functions generating}
\label{Dependent}
\begin{algorithmic}[1]
\REQUIRE $|\Omega|$, $n:$ number of mass functions, $d$ decision of another source
\FOR {$i=1$ to $n$}
	\STATE Find the set $S$ of all focal sets including $d$.
	\STATE Choose randomly $\mid F\mid$, the number of focal elements on $[1,| S |]$. 
	\STATE Choose randomly $\mid F\mid$ focal elements from $S$ noted $F$.
	\STATE Divvy the interval $[0,1]$ into $|F|$ continuous sub-intervals.
	\STATE Focal elements \textsc{bbm}s are intervals sizes.
\ENDFOR
\RETURN $n$ consistent mass functions
\end{algorithmic}
\end{algorithm}
\subsection{Results of tests}
Algorithms detailed in the previous section are used to test some cases of sources' dependence and independence. We note that in extreme cases where mass functions are certain or even when focal elements do not intersect; maximal values of independence are obtained. In the case of perfect dependence; mass functions have the same focal elements; however, clusters contain mass functions with consistent focal elements. Clustering is performed according to focal elements and clusters are perfectly linked.
\subsubsection{Independent sources}
In this paragraph, mass functions are independent. Focal elements and \textsc{bbm}s are randomly chosen ensuing algorithm \ref{Mass generating}. For tests, we choose $\mid\Omega\mid=5$ which is considered as medium-sized frame of discernment and $n=100$. Table \ref{tab} illustrates the mean of $100$ tests in the case of independent sources. The mean of $100$ tests for two dependent sources yields to a degree of independence $\gamma=0.68$, thus sources are independent. Assume that $m_1$ and $m_2$, given in table \ref{tabcomb}, are provided by two sources $s_1$ and $s_2$ which independence degree is given in table \ref{tab}. Combination of $m_1$ and $m_2$ is given in table \ref{expcomb}.\\
\indent To illustrate the case of three independent sources, three sets of $100$ independent mass functions are generated following algorithm \ref{Mass generating} with $\mid\Omega\mid=5$. The mean of $100$ tests are illustrated in table \ref{tab2}. 
\begin{table}[ht]
\begin{center}
\caption{Mean of $100$ tests on $100$ generated mass functions for two sources}
 \label{tab}
\begin{tabular}{|l|l|l|}
\hline
\scriptsize{\bfseries Dependence type}&\scriptsize{\bfseries Degree of independence} & \scriptsize{\bfseries Overall independence}\\
\hline
\multirow{2}{3cm}{Independence}
&$I_d(s_1,s_2)=0.68$, $\bar{I_d}(s_1,s_2)=0.32$&$\gamma=0.68$\\ 
  \cline{2-2}
&$I_d(s_2,s_1)=0.68$, $\bar{I_d}(s_2,s_1)=0.32$&\\ 
\hline 
\multirow{2}{3cm}{Dependence}
&$I_d(s_1,s_2)=0.34$, $\bar{I_d}(s_1,s_2)=0.66$&$\gamma=0.34$\\\cline{2-2}
&$I_d(s_2,s_1)=0.35$, $\bar{I_d}(s_2,s_1)=0.65$&\\
\hline
\end{tabular}
\end{center}
\end{table}
\begin{table}[ht]
\caption{Mixed combination of $m_1$ and $m_2$}
\label{expcomb}
\begin{center}
\begin{tabular}{| l | l | l | l | l | l | l |}
\hline
$2^\Omega$&$m_1$&$m_2$&$m_{Mixed}$&$m_{Mixed}$\\
&&&$\gamma=0.68$&$\gamma=0.34$\\
\hline
$\emptyset$&0&0&$0.092$&$0.076$\\
$a$&0.3&0.3&$0.3262$&$0.3881$\\
$b$&0&0&$0$&$0$\\
$a\cup b$&0&0&$0$&$0$\\
$c$&0.2&0&$0.1662$&$0.1531$\\
$a\cup c$&0.2&0.4&$0.2567$&$0.2583$\\
$b\cup c$&0&0&$0$&$0$\\
$a\cup b \cup c$&0.3&0.3&$0.1589$&$0.1244$\\
\hline
\end{tabular}
\end{center}
\end{table}
\begin{table}[ht]
\begin{center}
\caption{Mean of $100$ tests on $100$ generated mass functions for three independent sources}
 \label{tab2}
\begin{tabular}{|c|c|c|c|}
\hline
\scriptsize{\bfseries Sources}&\scriptsize{\bfseries Degree of independence} & \scriptsize{\bfseries Pairwise}& \scriptsize{\bfseries Overall}\\
&& \scriptsize{\bfseries independence}& \scriptsize{\bfseries independence}\\
\hline
\multirow{2}{2cm}{$s_1$-$s_2$}
&$I_d(s_1,s_2)=0.67$, $\bar{I_d}(s_1,s_2)=0.33$&$I(s_1,s_2)=0.67$&\\ 
  \cline{2-2}
&$I_d(s_2,s_1)=0.67$, $\bar{I_d}(s_2,s_1)=0.33$&&\\ 
\cline{1-3}
\multirow{2}{2cm}{$s_1$-$s_3$}
&$I_d(s_1,s_3)=0.68$, $\bar{I_d}(s_1,s_3)=0.32$&$I(s_1,s_3)=0.68$&$\gamma=0.68$\\ 
  \cline{2-2}
&$I_d(s_3,s_1)=0.68$, $\bar{I_d}(s_3,s_1)=0.32$&&\\ 
\cline{1-3}
\multirow{2}{2cm}{$s_2$-$s_3$}
&$I_d(s_2,s_3)=0.68$, $\bar{I_d}(s_2,s_3)=0.32$&$I(s_2,s_3)=0.68$&\\ 
  \cline{2-2}
&$I_d(s_3,s_2)=0.68$, $\bar{I_d}(s_3,s_2)=0.32$&&\\ 
\hline 
\end{tabular}
\end{center}
\end{table}
\subsubsection{Dependent sources}
In the case of dependent sources, mass functions are generated ensuing algorithms \ref{Consistent mass generating} and \ref{Dependent}. For tests, we choose $\mid\Omega\mid=5$ and $n=100$. We generate $100$ mass functions of both $s_1$ and $s_2$ for $100$ times and then compute the average of $\Id(s_1,s_2)$, $\Id(s_2,s_1)$ and $I(s_1,s_2)$. Table \ref{tab} illustrates the mean of $100$ independence degrees of two dependent sources providing each one $100$ randomly generated mass functions. These sources are dependent with a degree $1-\gamma=0.66$. In table \ref{expcomb}, $m_1$ and $m_2$ are combined using the mixed rule when $\gamma=0.34$.\\
To illustrate the case of three dependent sources, three sets of $100$ dependent mass functions are generated following algorithms  \ref{Consistent mass generating} and \ref{Dependent} when $\mid\Omega\mid=5$. The mean of $100$ degrees of independence are illustrated in table \ref{tab3}. 

Finally, assume that $m_1$, $m_2$ and $m_3$ of table \ref{comb2} are three mass functions defined on a frame of discernment $\Omega=\{a,b,c\}$ and provided by three dependent sources. The mixed combined mass function when their degree of independence is $\gamma=0.35$ is also given in table \ref{comb2}.
\begin{table}[ht]
\begin{center}
\caption{Mean of $100$ tests on $100$ generated mass functions for three dependent sources}
 \label{tab3}
\begin{tabular}{|c|c|c|c|}
\hline
\scriptsize{\bfseries Sources}&\scriptsize{\bfseries Degree of independence} & \scriptsize{\bfseries Pairwise}& \scriptsize{\bfseries Overall}\\
&& \scriptsize{\bfseries independence}& \scriptsize{\bfseries independence}\\
\hline
\multirow{2}{2cm}{$s_1$-$s_2$}
&$I_d(s_1,s_2)=0.35$, $\bar{I_d}(s_1,s_2)=0.65$&$I(s_1,s_2)=0.34$&\\ 
  \cline{2-2}
&$I_d(s_2,s_1)=0.34$, $\bar{I_d}(s_2,s_1)=0.66$&&\\ 
\cline{1-3}
\multirow{2}{2cm}{$s_1$-$s_3$}
&$I_d(s_1,s_3)=0.32$, $\bar{I_d}(s_1,s_3)=0.68$&$I(s_1,s_3)=0.31$&$\gamma=0.35$\\ 
  \cline{2-2}
&$I_d(s_3,s_1)=0.31$, $\bar{I_d}(s_3,s_1)=0.69$&&\\ 
\cline{1-3}
\multirow{2}{2cm}{$s_2$-$s_3$}
&$I_d(s_2,s_3)=0.36$, $\bar{I_d}(s_2,s_3)=0.64$&$I(s_2,s_3)=0.35$&\\ 
  \cline{2-2}
&$I_d(s_3,s_2)=0.35$, $\bar{I_d}(s_3,s_2)=0.65$&&\\ 
\hline 
\end{tabular}
\end{center}
\end{table}
\begin{table}[ht]
\caption{Mixed combination of $m_1$, $m_2$ and $m_3$}
\label{comb2}
\begin{center}
\begin{tabular}{| l | l | l | l | l | l | l |}
\hline
$2^\Omega$&$m_1$&$m_2$&$m_3$&$m_{Mixed}$\\
&&&&$\gamma=0.35$\\
\hline
$\emptyset$&0&0&0&0\\
$a$&0&0&0&0\\
$b$&0&0&$0$&$0$\\
$a\cup b$&0&0&$0$&$0$\\
$c$&0.03&$0.05$&$0$&$0.32$\\
$a\cup c$&0.39&0.07&$0.04$&$0.24$\\
$b\cup c$&$0.3$&$0.47$&$0.22$&$0.29$\\
$a\cup b \cup c$&0.28&0.41&$0.74$&$0.15$\\
\hline
\end{tabular}
\end{center}
\end{table}

\section{Conclusion}
In this paper, we proposed a method to learn sources cognitive independence in order to use the appropriate combination rule either when sources are cognitively dependent or independent.
Sources are cognitively independent if they are different; not communicating and they have distinct evidential corpora. The proposed statistical approach is based on a clustering algorithm applied to mass functions provided by several sources. A pair of sources independence is deduced from weights of linked clusters after a matching of their clusters.
Independence degree of sources can either guide the choice of the combination rule if it is either $1$ or $0$; when it is a degree over $]0,1[$, we propose a new combination rule that weights the conjunctive and cautious combinations with sources' independence degree.
\bibliographystyle{elsarticle-num}
\bibliography{biblio-extracted-VF}








\end{document}